\pdfoutput=1

\documentclass[11pt]{article}

\usepackage[preprint]{acl}

\usepackage{times}
\usepackage{latexsym}
\usepackage[T1]{fontenc}
\usepackage[utf8]{inputenc}

\usepackage{microtype}
\usepackage{inconsolata}

\usepackage{graphicx}
\usepackage{lipsum}
\usepackage[nolist]{acronym}
\usepackage[inline]{enumitem}
\usepackage{amsmath}
\usepackage[colorinlistoftodos]{todonotes}
\usepackage{algorithm}
\usepackage{algpseudocode}
\usepackage{amsfonts}
\usepackage{amssymb}
\usepackage{tabularx}
\usepackage{multirow}
\usepackage{lscape}
\newcolumntype{B}{>{\hsize=1.45\hsize}X}
\newcolumntype{S}{>{\hsize=.55\hsize}X}

\title{SeCoKD: Aligning Large Language Models for In-Context Learning with Fewer Shots}

\author{Weixing Wang,  Christoph Meinel,  Haojin Yang \\
  Hasso Plattner Institute, University of Potsdam, Germany \\
  \texttt{\{weixing.wang, christoph.meinel, haojin.yang\}@hpi.de} \\}

\begin{document}
\maketitle
\begin{abstract}
Previous studies have shown that demonstrations can significantly help \ac{LLMs} perform better on the given tasks. However, this so-called \ac{ICL} ability is very sensitive to the presenting context, and often dozens of demonstrations are needed. In this work, we investigate if we can reduce the shot number while still maintaining a competitive performance. We present \textit{SeCoKD}, a self-\ac{KD} training framework that aligns the student model with a heavily prompted variation, thereby increasing the utilization of a single demonstration. We experiment with the \textit{SeCoKD} across three \ac{LLMs} and six benchmarks focusing mainly on reasoning tasks. Results show that our method outperforms the base model and \ac{SFT}, especially in zero-shot and one-shot settings by 30\% and 10\%, respectively. Moreover, SeCoKD brings little negative artifacts when evaluated on new tasks, which is more robust than \acl{SFT}.
\end{abstract}

\section{Introduction}
When scaling up \ac{LLM}s, the ability of \ac{ICL} emerges \citep{brown2020language, agarwal2024many, dong2022survey}. Models can learn from a few demonstrations and thus can be generalized to various downstream tasks without updating the parameters \citep{wei2023larger}. However, the mechanism behind the few-shot learning ability remains unclear. Large language models are very sensitive to the quality of demonstrations, such as the number of demonstrations \citep{pan2023context, chen2023many}, the order of reasoning steps \citep{lu2021fantastically, zhao2021calibrate}, and the correctness of labels \citep{halawi2023overthinking}. Moreover, the design of a demonstration also plays an important role \citep{zhao2021calibrate, wang2022towards, fu2022complexity, wei2022chain}. As a result, it is not trivial to design a proper demonstration and the importance of prompt engineering continues to increase \citep{reynolds2021prompt, dong2022survey, zhou2022least}. Currently, it is common to use dozens of demonstrations together to overcome the possible weakness of a single prompt. However, we argue that humans often do not need more than two examples in the context of Q\&A. One demonstration can serve as a guideline and show the correct format for answering the question, but more similar demonstrations are irrelevant to the correctness of the answer. In other words, humans do one-shot or zero-shot learning and they are not few-shot learners. 

In this paper, we propose a simple yet effective \ac{KD} method called \textit{SeCoKD}, which stands for \textbf{Se}lf \textbf{Co}ntext \textbf{K}nowledge \textbf{D}istillation. Our method significantly reduces the number of demonstrations needed in the context by increasing the utilization of a single demonstration. The intuition is that since an \ac{LLM} can answer a question correctly when triggered by a certain amount of external information (few-shot learning), we could use less information (one-shot learning) by aligning the model space and the task space through self-\ac{KD}. It differs from internalizing knowledge; instead, it promotes the model to utilize existing information to activate its internal knowledge, a process previously achieved by providing a handful of examples. 

\begin{figure*}[t]
  \includegraphics[width=0.98\linewidth]{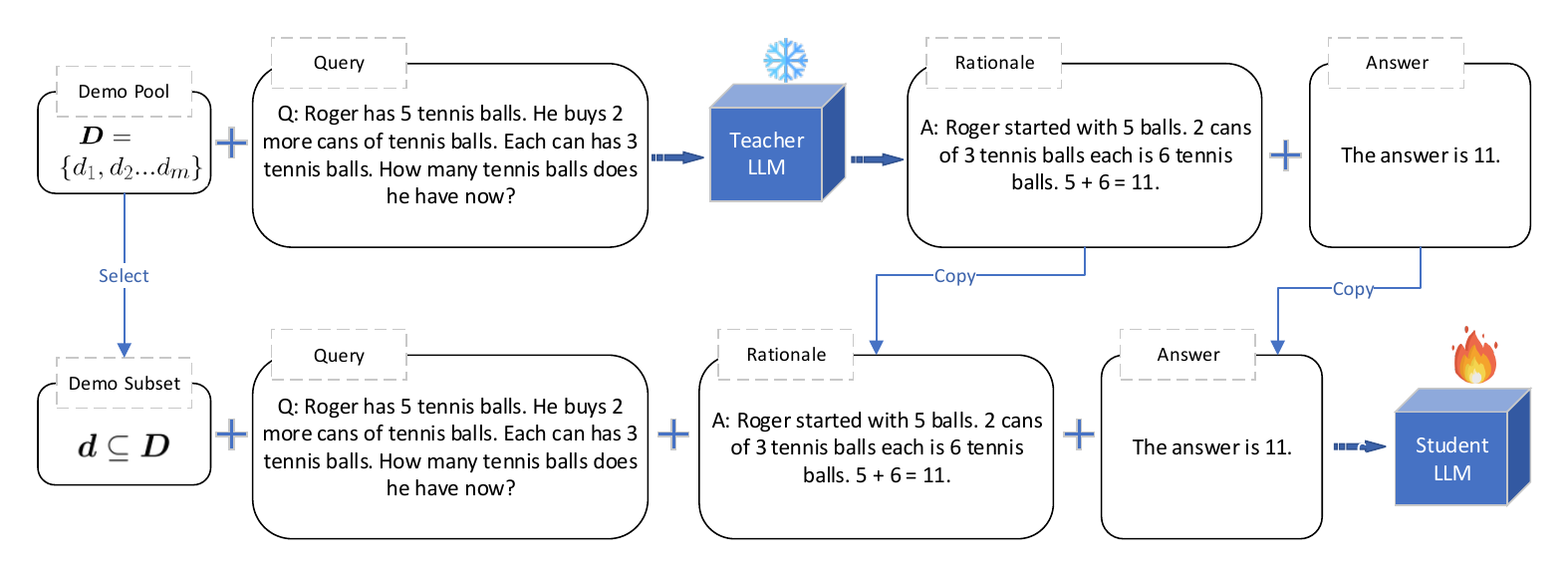}
  \caption {Overview of the SeCoKD framework. The teacher model first generates high-quality rationale and answers for a query through 8-shot \ac{ICL}. Then a student is trained using fewer demonstrations and the teacher's output.  }
  \label{fig:model}
\end{figure*}

First, we show that SeCoKD strongly improves the model performance on zero-shot and one-shot learning. We also consider the model trained with supervised fine-tuning as a strong baseline. In comparison, our method achieves better performance, especially when the original training set doesn't provide reasoning steps. For example, when performing one-shot \ac{ICL} on the ARC-C \citep{Clark2018ThinkYH} dataset, the Mistral-7B fine-tuned with our method scores 60\% accuracy, 10\% higher than the initial model and 3\% higher than the \ac{SFT} version.

Second, we demonstrate that SeCoKD not only enhances performance on the training task but also maintains robustness across different tasks, unlike \ac{SFT}, which can reduce accuracy on unseen tasks. This indicates that our method is more robust compared to \ac{SFT}.

Third, empirical experiments suggest that SeCoKD simplifies tasks by converting difficult queries into easier ones when the same demonstration is provided. In contrast, while \ac{SFT} occasionally outperforms SeCoKD in accuracy, its improvements are inconsistent: some queries that are initially easy for the base model become significantly more challenging after \ac{SFT}.

In summary, the contributions of this study are as follows:
\begin{itemize}
  \item To the best of our knowledge, this work represents the first approach deliberately designed to reduce the number of demonstrations used for \ac{ICL} by enhancing the model's ability to utilize a single demonstration.
  \item We design a \ac{KD} training pipeline called SeCoKD and conduct comprehensive empirical experiments on various reasoning tasks in the \ac{ICL} setting. In total, 6 datasets and 3 different models are used in this study.
  \item We investigate the robustness of SeCoKD in comparison to the \ac{SFT} and show that our method not only provides more consistent improvements but also generalizes well to unseen tasks.
\end{itemize}

\section{Related Work}

\subsection{Few-Shot In-Context Learning}
Recent work \citep{radford2019language} demonstrated that large \acl{PLM}s can perform incredibly well on standard NLP tasks without being fine-tuned on task-specific datasets. Furthermore, \citet{brown2020language} suggested that the performance can be improved by feeding extra information in the input context. It is typically done by providing \textit{demonstrations} of the same task. A \textit{demonstration} refers to a text sequence that contains at minimum a query and its corresponding answer, concatenated by a predefined pattern. Additional information such as instructions and rationale can also be included. Although being competitive in certain tasks, \ac{ICL} suffers from instability. Its performance depends heavily on the model size \citep{wei2023larger}, the overall format of sequences \citep{min2022rethinking}, number of demonstrations \citep{chen2023many, halawi2023overthinking}, etc. As a result, there are no gold standards for designing context and the studies about \ac{ICL} are mostly empirical.

On one hand, some works showed that enriching context can be beneficial. \citet{agarwal2024many} proposed many-shot learning to make full use of the allowed context length. With hundreds or thousands of demonstrations, models constantly perform better than just using a few demonstrations. 

On the other hand, \citet{chen2023many} revealed that more demonstrations do not always bring benefits. Instead, \ac{ICL} with one proper demonstration may perform better than few-shot learning using multiple random demonstrations. Towards more efficient \ac{ICL}, existing works focus on demonstration selection \citep{li2023finding, wu2022self, li2023unified} or context compression \citep{wingate2022prompt, ge2023context}. \citet{zhang2022active} proposed a reinforcement learning approach to select a handful of demonstrations from up to 1000 examples. \citet{pan2024llmlingua} developed a task-agnostic prompt compression technique that achieved a compression ratio of up to 5x without losing much performance. However, there is no existing approach to improve the model's internal ability to handle arbitrary demonstration, which can lead to a more fundamental solution. To fill this research void, we focus on reducing the number of demonstrations as much as possible while maintaining performance and robustness. 

\subsection{Distillation of Language Models}
\acl{KD} \citep{hinton2015distilling, gou2021knowledge} is a technique in machine learning that involves transferring knowledge from a larger, more complex model (often referred to as the "teacher" model) to a smaller, more efficient model (known as the "student" model). The goal is to enable the student model to achieve performance similar to or close to that of the teacher model but with reduced computational cost and lower resource requirements. \citet{xu2024survey} recently summarized three main motivations for applying \ac{KD} in context of \ac{LLM}: 
\begin {enumerate*} [label=\itshape\alph*\upshape)]
\item trying to let the open-source models mimic and learn from the more powerful closed-source model, 
\item offering compressed and efficient models,
\item enhancing models using self-generated data through self \ac{KD}. 
\end {enumerate*}

The last point is an emerging research topic since the recent \ac{LLM}s can generate high-quality data that can be used for self-improvement. In \citeposs{sun2024principle} work, the authors synthesized around 360k training samples with LLaMA-65b and later fine-tuned the same model with these data. Thanks to the self-alignment between the model and the generated data, their model surpassed many models trained with human-curated samples. Extending the idea of self-improvement, we propose to use the same model to generate a high-quality rationale for a query that serves as the most aligned supervision to train a student model.

\section{SeCoKD Overview}
The primary training objective of SeCoKD is to have the student model emulate the teacher model, which is activated by a handful of demonstrations. Concretely, let $\boldsymbol{D} = \left \{ d_1,d_2,d_3...d_n \right \}$ denotes a set of demonstrations and $\boldsymbol{d}\subseteq \boldsymbol{D}$ denotes a subset. $\left ( x,y,\theta  \right )$ are the input query, true label, and model parameters, respectively. In the setting of few-shot learning we have $P_{\mathcal{M}}=\left ( y\mid x,\boldsymbol{D},\theta_{\mathcal{M}}  \right )$ for the model $\mathcal{M}$. After applying our training method, we showcase that the updated Model ${\mathcal{M}}'$ also performs well with a high $P_{{\mathcal{M}}'}=\left ( y\mid x,\boldsymbol{d},\theta_{{\mathcal{M}}'}  \right )$. Since we focus on a self-distillation manner and we fine-tune the model with LoRA, the expression can be rewritten as $P_{{\mathcal{M}}'}=\left ( y\mid x,\boldsymbol{d},\theta_{LoRA}, \theta_{\mathcal{M}}  \right )$.
As depicted in Figure~\ref{fig:model}, the whole pipeline can be divided into two steps. 

First, the teacher model is prompted with a set of demonstrations (demonstration pool) and a query. Each demonstration contains a question, a rationale, and an answer. The creation of the demonstration pool is detailed in~\ref{sec:demonstration}. The reasons to include some reasoning steps are two-fold:
\begin {enumerate*} [label=\itshape\alph*\upshape)]
\item It is shown that \ac{CoT} prompting increases the reasoning ability of \ac{LLM}s and thus the performance will be better \citep{wei2022chain, shao2023synthetic}.
\item We need more tokens generated from the teacher model as supervision of the student model.
\end {enumerate*}
For each task, we use a carefully curated demonstration set as gold samples. We then extract the reasoning part and the answer from the teacher model's output and save them for later use.

Second, we randomly sample a subset of the available demonstrations, concatenate it with the same query as in the first step, and use this sequence as input for the student model. Then we apply Sequential-Level \ac{KD} \citep{kim2016sequence} to fine-tune the student model.

To explain the whole pipeline mathematically, we first obtain the teacher output as
\begin{equation}
    \textbf{r} =g(f_{teacher} \left ( \boldsymbol{D},x \right ))
\end{equation}
where $f(\cdot)$ is the generation function and $\boldsymbol{D}$ is the demonstration pool. We use the extraction function $g(\cdot)$ to obtain the teacher-forcing supervision for the student model.
Our training objective is to find parameters $\theta$ of the student model $S$ that maximize the sequential-level log-probability sum:
\begin{equation}
\begin{aligned}
\mathcal{M}_{Seq}(\theta) & =\mathbb{E}_{(\boldsymbol{pre}, \boldsymbol{r}) \sim \mathcal{D}}\left[\log S_\theta(\hat{\boldsymbol{r}}=\boldsymbol{r} ; \boldsymbol{pre})\right] \\
& =\mathbb{E}_{\mathcal{D}}\left[\sum_{i=1}^{L_{r}} \log S_\theta\left(\hat{r}_i=r_i ; \boldsymbol{pre}, \boldsymbol{r}_{<i}\right)\right]
\end{aligned}
\end{equation}
where $\boldsymbol{pre}$ denotes the student input, containing the selected demonstrations and the query.
Given this objective, the corresponding loss function to be minimized can be framed as:
\begin{equation}
    \mathcal{L}(\theta)=
    -\frac{1}{N} \sum_{j=1}^N \sum_{i=1}^{L_{\text {r }}^{(j)}} \log S_\theta\left(\hat{r}_i^{(j)}=r_i^{(j)} \mid p^{(j)}, r_{<i}^{(j)}\right)
\end{equation}

\paragraph{Demonstration Selection}
We develop two strategies to select demonstrations for the student model. \textbf{SeCoKD-S} randomly samples one demonstration out of the demonstration pool. This represents the extreme case where we hypothesize that one example can already provide enough guidance for the model. \textbf{SeCoKD-M}, on the other hand, samples a different number of demonstrations from 1\textasciitilde4 for the student model, providing a stronger initial guidance.

\section{Experimental Settings}
We aim to evaluate the performance of SeCoKD compared to directly supervised fine-tuning and the base model. Inspired by \citet{wei2022chain}, we choose 6 popular benchmarks, covering topics of arithmetic reasoning, commonsense reasoning, and symbolic reasoning. We conducted experiments with some of the most advancing \ac{LLM}s. However, we could only test the models with less than 10 billion parameters due to the computation limits. Our code is publicly available on \href{https://github.com/weixingW/secokd-icl.git}{Github}.

\subsection{LLMs}
We evaluate our method on three GPT-like auto-regressive transformer language models. We use the 4-bit quantized version to save computation resources \citep{dettmers2022llmint8}. The \textbf{Llama 2-7B} \citep{touvron2023llama} is one of the most popular open-source \ac{LLM}s. \textbf{Llama 3-8B} \citep{llama3modelcard} is the latest member in the Llama family and appears to be the SOTA in various benchmarks. We also use the \textbf{Mistral-7B} \citep{jiang2023mistral} which leverages the sliding window attention (SWA) mechanism to handle variants sequence lengths effectively. We conducted all experiments on a single NVIDIA V100 (40G) GPU. For the training, we use the same LoRA\footnote{\url{https://huggingface.co/docs/diffusers/en/training/lora}} \citep{hu2021lora} configuration for all models, and the trainable parameters thus reduce to around 0.18\% of the full size. All results reported are the average of three runs and training-related hyperparameters are listed in~\ref{sec:hyperparameters}.
\subsection{Datasets}


\begin{figure}[t]
  \includegraphics[width = \columnwidth]{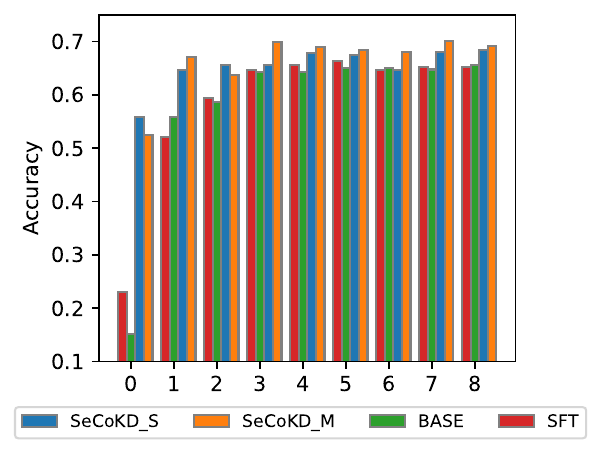}
  \caption {Comparison of 4 methods with different shot numbers. The X-axis represents the number of demonstrations used for inference. The Y-axis shows the average accuracy of all six tasks. SeCoKD significantly outperforms the other two baselines in zero-shot and one-shot scenarios.}
  \label{fig:exp1_bar}
\end{figure}

We evaluate all methods on 6 datasets covering 3 types of reasoning tasks. For mathematical reasoning, we selected three datasets. The \textbf{GSM8K} \citep{cobbe2021training} contains 8.5K high-quality and diverse text-based grade school math problems. The \textbf{SVAMP} \citep{patel2021nlp} applies different types of variations to the existing math problems and creates a more robust benchmark. In \textbf{AQUA-RAT} \citep{ling2017program}, the answers to the math problems are multiple choices. This introduces diversity into our experiments. For the commonsense reasoning tasks, we selected \textbf{ARC-C} \citep{clark2018think} which contains relatively difficult natural grade-school level questions, and the \textbf{CSQA} \citep{talmor-etal-2019-commonsenseqa} which utilized crowd-workers to create multiple-choice questions that cover a wide range of topics. We chose the \textbf{Coin-Flip} dataset introduced by \citet{wei2022chain} for the symbolic reasoning task. In this task, the model is asked if a coin is still heads-up after $n$ people flip it. For each task, we randomly sample 800 pieces of data for training and 200 for testing.

\subsection{Training SFT model}
The key of SeCoKD is to use teacher-generated rationale to align the target task with the student model. As a comparison, we train a separate model with normal SFT to exclude teacher supervision by replacing teacher-generated supervision with standalone rationales that either can be found in the dataset or created. For more details please refer to~\ref{sec:rationale}. Specifically, we follow the step 2 of training SeCoKD with the following changes: \begin {enumerate*} [label=\itshape\alph*\upshape)]
\item For demonstration we use the whole demonstration pool $\textbf{D}$ instead of a subset of it.
\item We use standalone rationales and answers instead of generated ones.
\end {enumerate*}
With such a training schema we guarantee that the SFT model uses the same amount of data as SeCoKD and thus isolates the effect of the knowledge distillation.

\section{Results and Discussion}

For inference and evaluation, we test the zero-shot and few-shot learning ability by using 0 to 8 demonstrations in the same demonstration pool.

\begin{figure}[htbp]
  \centering
  \includegraphics[width=\columnwidth]{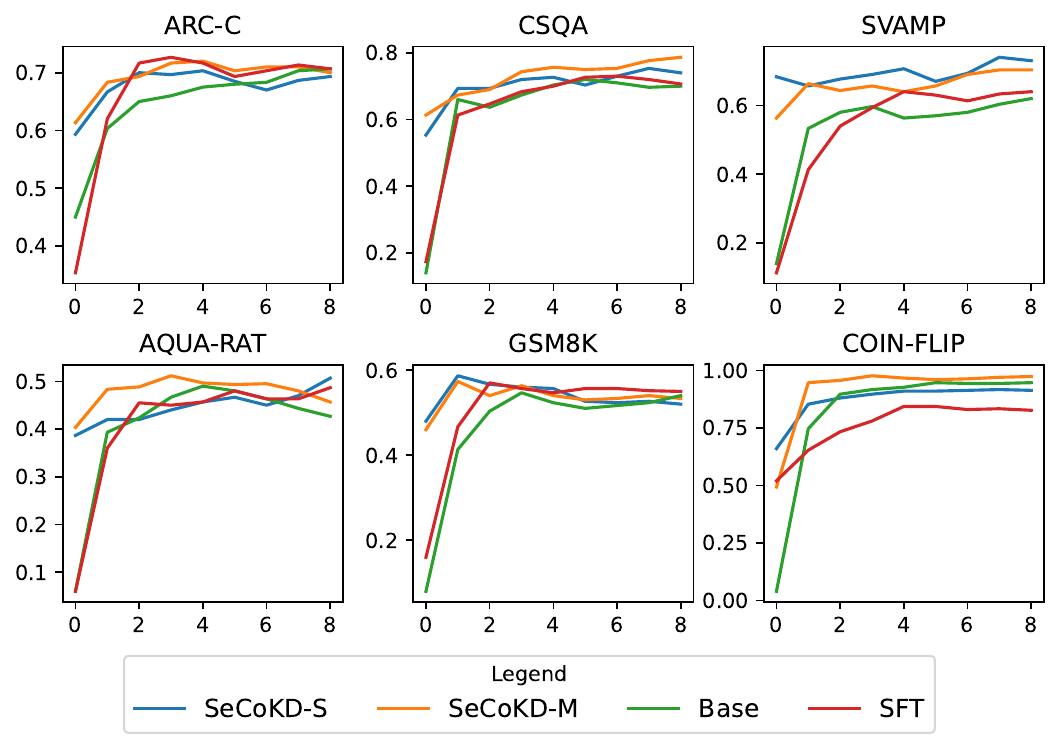}
  \caption {Few-Shot performance on each task. The X-axis represents the number of demonstrations used for inference. Our methods SeCoKD-S and SeCoKD-M perform much better in zero-shot and one-shot compared to the two baselines.}
  \label{fig:exp1_line}
\end{figure}
\subsection{Results for Few-Shot ICL}

Figure~\ref{fig:exp1_bar} shows an overall comparison between SeCoKD and baseline methods. The underlying model here is Llama 3-8B and more results regarding different LLM structures can be found in~\ref{sec:results}. First, while all fine-tuned models perform better than the base model, the two variants SeCoKD-S and SeCoKD-M are better than SFT in most scenarios. We observe the largest margins in the zero-shot case, meaning that context information is successfully compressed. Second, we notice that the difference between the SeCoKD-S and SeCoKD-M is quite small. This means that in the distillation process, the model does not need a strong initial context to align with the guidance from the teacher. In the following experiments, we only use the SeCoKD-S to reduce computational resources. Last, starting from Four-Shot, adding more demonstrations seems to have no more positive impact on the performance of all methods. This observation is consistent with the study from \citet{min2021metaicl}. In their work, \ac{ICL} brings only marginal improvements also after around 4 demonstrations. This indicates that there is a performance upper bond for the model that can be lifted by training, not by \ac{ICL}.

In Figure~\ref{fig:exp1_line} we look separately at performance comparison on each dataset. We could see that in all tasks, the base model struggles in the zero-shot case, delivering the poorest performance. However, when providing more demonstrations, the performance is significantly increased up to an upper bound. After that, more demonstrations seem to have limited help, sometimes even degrading the performance for example for the AQUA-RAT task. The models trained with \ac{SFT} also perform generally not well in the zero-shot settings except for the COIN-FLIP task.  It even degrades the model's performance on ARC-C. When providing more demonstrations, \ac{SFT} can offer only limited improvement. Conversely, models trained with SeCoKD exhibit significantly better zero-shot performance across all tasks. Furthermore, the one-shot accuracy with SeCoKD already achieves optimal performance, indicating that more than one demonstration is unnecessary due to the effectiveness of the \ac{KD} pipeline. 

\begin{table*}[htbp]
\begin{tabular}{c|ccccccc}
\hline
                                                                       &          & \textbf{ARC-C} & \textbf{CSQA} & \textbf{SVAMP} & \textbf{AQUA-RAT} & \textbf{GSM8K} & \textbf{COIN-FLIP} \\ \hline \hline
\multirow{4}{*}{\begin{tabular}[c]{@{}c@{}}Llama 3\\ -8B\end{tabular}} & Base     & 0.6            & 0.66          & 0.53           & 0.39          & 0.41           & 0.74               \\ \cline{2-8} 
                                                                       & SFT      & 0.64           & 0.62          & 0.49           & 0.38          & 0.48           & 0.75               \\ \cline{2-8} 
                                                                       & SeCoKD-S & 0.67           & \textbf{0.69} & \textbf{0.66}  & 0.44          & \textbf{0.58}  & 0.85               \\ \cline{2-8} 
                                                                       & SeCoKD-M & \textbf{0.68}  & 0.67          & \textbf{0.66}  & \textbf{0.48} & 0.57           & \textbf{0.94}      \\ \hline \hline
\multirow{4}{*}{\begin{tabular}[c]{@{}c@{}}Llama 2\\ -7B\end{tabular}} & Base     & 0.4            & 0.42          & 0.29           & 0.14          & 0.05           & 0.51                \\ \cline{2-8} 
                                                                       & SFT      & 0.41           & 0.45          & 0.27           & \textbf{0.19} & 0.12           &  0.58                  \\ \cline{2-8} 
                                                                       & SeCoKD-S & \textbf{0.48}  & 0.52          & 0.3            & 0.15          & \textbf{0.19}  &  0.62                  \\ \cline{2-8} 
                                                                       & SeCoKD-M & 0.45           & \textbf{0.53} & \textbf{0.32}  & 0.14          & 0.18           &\textbf{0.63}                    \\ \hline \hline
\multirow{4}{*}{\begin{tabular}[c]{@{}c@{}}Mistral\\ -7B\end{tabular}} & Base     & 0.5            & 0.68          & 0.53           & 0.25          & 0.28           & 0.61                 \\ \cline{2-8} 
                                                                       & SFT      & 0.57           & \textbf{0.71}  & 0.64           & \textbf{0.33} & 0.49           & 0.63                   \\ \cline{2-8} 
                                                                       & SeCoKD-S & 0.59           & 0.68          & 0.62           & 0.27          & \textbf{0.6}   & 0.74                   \\ \cline{2-8} 
                                                                       & SeCoKD-M & \textbf{0.60}  & 0.69          & \textbf{0.65}  & 0.28          & 0.58           &  \textbf{0.78}                  \\ \hline
\end{tabular}
\caption{Comparison of one-shot accuracy on different tasks and different models. Bold values represent the best results within a model structure. We could see that in most cases SeCoKD performs the best. }
\label{tab:exp1}
\end{table*}
Table~\ref{tab:exp1} presents a comparison of one-shot accuracy on six different tasks across three models: Llama 3-8B, Llama 2-7B, and Mistral-7B. The methods compared are Base, SFT, SeCoKD-S, and SeCoKD-M. SeCoKD generally performs best across different tasks and models, showing the highest accuracy in most cases. For instance, in the ARC-C task, SeCoKD-M achieves 68\% accuracy with the Llama 3 model, outperforming the Base method at 60\% and SFT at 62\%. Similarly, in the GSM8K task, SeCoKD-M reaches 60\% accuracy with the Mistral model, while Base and SFT score 28\% and 44\%, respectively. SeCoKD-S often closely follows SeCoKD-M or performs slightly better, such as in the CSQA task with Llama 3, where SeCoKD-S scores 69\% compared to SeCoKD-M's 67\%. In contrast, Base and SFT methods typically show lower performance compared to SeCoKD methods, with SFT sometimes even performing worse than the Base model, especially in the Llama 2 model, where SFT scores 41\% in the CSQA task compared to the Base's 42\%. Overall, SeCoKD methods, significantly improve one-shot accuracy across various tasks compared to Base and SFT methods, especially in more complex models like Llama 3 and Mistral.

\subsection{Robustness of SeCoKD}

We demonstrate the superiority of SeCoKD over SFT by highlighting its robustness in cross-task testing scenarios. Our approach involves tuning a model on each individual task and then evaluating it not only on the test set of the same task but also on the test sets of other tasks. The rationale behind this experiment is twofold:
\begin{enumerate*}
\item A model that is effectively trained on a specific task should exhibit the best performance on that task compared to other model variants.
\item The training objective aims to enhance the model's ability to utilize demonstrations. Therefore, ideally, training on one task should also positively impact the model's performance on other tasks.
\end{enumerate*}

\begin{figure}[htbp]
    \centering
    \includegraphics[width=\columnwidth]{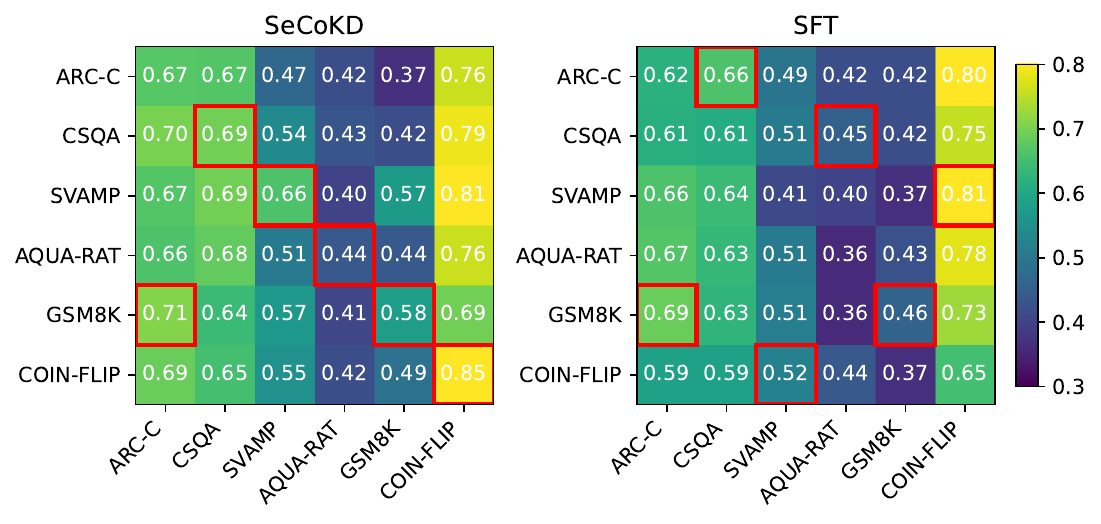}
    \caption {Cross-task tests of one-shot performance on different benchmarks. The Y-axis is the training task, and the X-axis represents the testing task. The cell value represents the absolute accuracy and we use the red boxes to highlight the best score in a column. For example, the top right cell shows the evaluation accuracy on the COIN-FLIP task when the model is trained on the ARC-C task.}
    \label{fig:exp2_heat}
\end{figure}

Figure~\ref{fig:exp2_heat} shows the accuracy of the Llama 3 model on different tasks. When comparing within a column, it is evident that SeCoKD generally achieves the highest accuracy on the task used for training, with the exception of the commonsense reasoning task ARC-C. Here the model trained with the mathematical reasoning task GSM8K performs the best, 4\% better than the model trained with ARC-C task. In this case, the model trained on the mathematical reasoning task GSM8K outperforms the one trained on ARC-C by 4\%. However, training with SFT does not yield the best results for the specific task in most cases. For instance, in the AQUA-RAT evaluation, the model trained on CSQA performs nearly 10\% better than the one trained on AQUA-RAT. For the COIN-FLIP task, this performance gap can reach up to 15\%.

\begin{figure}
    \centering
    \includegraphics[width=\columnwidth]{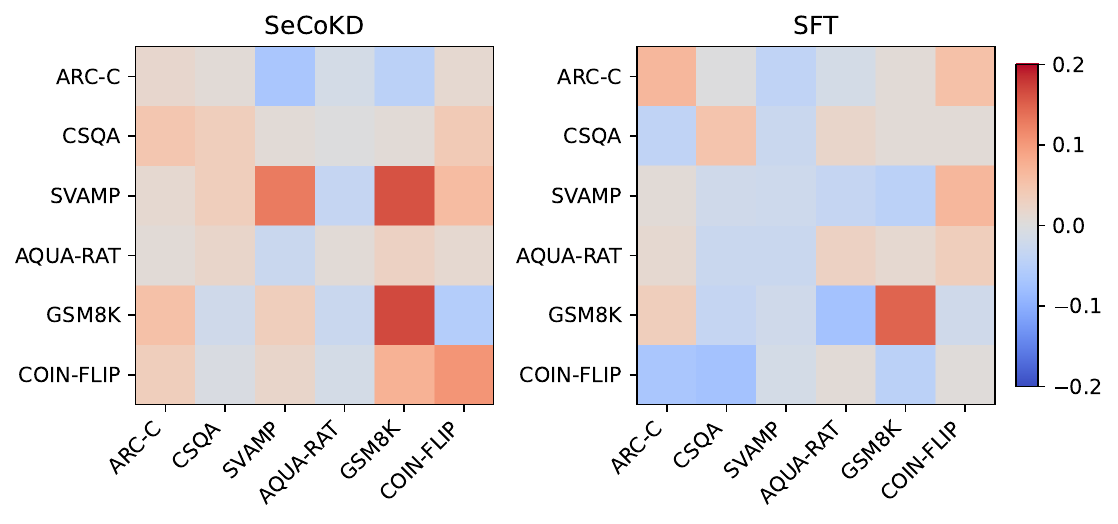}
    \caption{Cross-task evaluation of one-shot performance across different benchmarks. The Y-axis indicates the training task, while the X-axis represents the testing task. To assess the impact of the training method on model performance, we subtract the baseline accuracy from the accuracy achieved post-training. A red cell color indicates that the trained model outperforms the base model, whereas blue cells signify a decline in performance after training.}
    \label{fig:exp2_heat_base}
\end{figure}

We also utilize cross-task testing to showcase the high robustness of SeCoKD by visualizing the performance gap between post-training and baseline models in Figure~\ref{fig:exp2_heat_base}. The color scale indicates the change in post-training compared to the baseline in terms of one-shot accuracy, with red indicating improvement and blue indicating a decline. We can see that SeCoKD has a more significant positive transfer effect, as evidenced by the broader spread of red cells across the heatmap, suggesting it generalizes better across tasks compared to SFT.

\subsection{Simplifying tasks with SeCoKD}

In this section, we emphasize the benefits of SeCoKD by showing that training with this method makes a task easier to solve. Inspired by \citet{chen2023many}, we also provide a metric to make the measurement of easiness more tangible. 
\paragraph{\textbf{positive} and \textbf{negative} demonstration}
Following the definitions in Chen's paper, a positive demonstration helps the model to answer correctly in the setting of one-shot learning. A negative demonstration, in contrast, results in a false answer. 
\paragraph{\textbf{Easy}, \textbf{Hard}, and $\textbf{Hard}\ast$ sample}
For each task, there are in total eight existing or hand-crafted gold demonstrations. We conduct one-shot experiments using these demos and classify the sample into three categories based on the number of positive demonstrations $n$: 
\begin{itemize}
\item easy: $n \geqslant 6$. 
\item hard: $6> n >  1$.
\item $hard\ast $: $1\geq n$.
\end{itemize}

\begin{figure}
    \centering
    \includegraphics[width=\columnwidth]{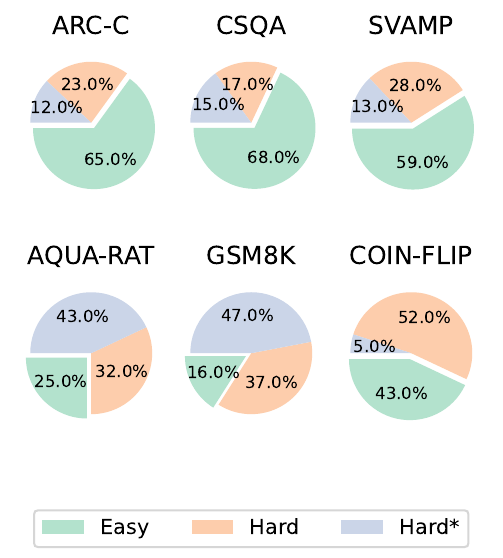}
    \caption{Queries in a dataset are categorized into three classes, representing their easiness to be solved with one-shot \ac{ICL}. $Hard\ast$ means none of the existing demonstrations can lead to a correct answer. We can see that all datasets are very biased.}
    \label{fig:exp3_base}
\end{figure}

In the following experiments, we focus on the Mistral model, the conclusions drawn from the other two models are similar. Figure~\ref{fig:exp3_base} visualizes the initial category distribution. We can see that the AQUA-RAT dataset stands out with a large portion of Hard* tasks (43\%), indicating that it is predominantly challenging. Only a quarter of the dataset is categorized as Easy. GSM8K is also highly challenging with the smallest Easy category (16\%) and a majority of samples falling under Hard (37\%) and Hard* (47\%), highlighting the dataset's complexity. As a result, we could observe very low one-shot accuracy for these two datasets in Table~\ref{tab:exp1}, both below 30\%.  The majority of the data in ARC-C, SVAMP, and CSQA is classified as Easy, suggesting that a significant portion of the samples can be easily addressed using demonstrations in one-shot learning. However, there is still a notable portion that ranges from hard to very hard, indicating a substantial amount of more challenging tasks. 

\begin{table*}[htpb]
\centering
\begin{tabular}{ccccccc}
\hline
       & ARC-C        & CSQA          & SVAMP         & AQUA-RAT      & GSM8K          & COIN-FLIP   \\ \hline
SeCoKD    & \textbf{1.22} & \textbf{0.96} & \textbf{2.18} & \textbf{1.35} & \textbf{3.56} & \textbf{4.57} \\ \hline
SFT & 0.74        & 0.46         & 1.53         & 1.21          & 2.12             & 1.78       \\ \hline
\end{tabular}
\caption{Improvement Scores of SFT and SeCoKD. Larger is better.}
\label{tab:exp3_tab}
\end{table*}

\paragraph{Improvement Score}
To measure the change in data distribution with regard to the three categories, we develop a metric called \textit{improvement score (IS)}:
\begin{equation}
IS=\exp \left ( \frac{1}{N} \sum_{i=0}^{N} \frac{(n_i-m_i)}{D}\right )
\end{equation}
where $n$ and $m$ represent the number of positive demonstrations obtained using the fine-tuned model and the base model, respectively. $D$ is the size of the demonstration set which is 8 in our case. A higher IS value indicates that more demonstrations are considered positive for a given query, making the query an easier task. This metric is advantageous because it evaluates the transformation of individual samples into easier ones, rather than just comparing the overall data distribution. Essentially, IS measures the proportion of samples that become easier to handle, offering a more nuanced assessment of the training method. 

From Table~\ref{tab:exp3_tab}, it is evident that SeCoKD consistently outperforms SFT across all datasets. SeCoKD demonstrates particularly high scores in the COIN-FLIP and GSM8K datasets. While SFT occasionally achieves better accuracy scores, it often leads to a significant portion of previously easy tasks becoming more difficult, which is an undesirable outcome. For instance, as shown in Table~\ref{tab:exp1}, the Mistral model trained with SFT achieves an accuracy score of 0.32 on the AQUA-RAT dataset, whereas SeCoKD scores slightly lower at 0.28. However, SFT has an Improvement Score (IS) of 1.01, which is smaller than the IS achieved by the SeCoKD method. This indicates that despite the higher accuracy, SFT makes the tasks more challenging overall.
SeCoKD, on the other hand, excels at preserving previously positive tasks while effectively converting difficult tasks.

\section{Conclusion}
We introduce SeCoKD, a \acl{KD} framework that enhances the \acl{ICL} abilities of \acl{LLMs} using fewer demonstrations. Our experiments show that SeCoKD significantly improves model performance, robustness, and efficiency compared to traditional methods like \acl{SFT}.

SeCoKD-trained models excel with minimal demonstrations, achieving optimal accuracy with just one demonstration. They outperform base models by an average of 10\% in one-shot ICL scenarios and show enhanced robustness without negative cross-task performance impacts, which is a common issue with SFT.
Cross-task testing highlights SeCoKD's robustness and generalization, with models performing well not only on their training tasks but also on other tasks. This indicates effective compression and alignment of task-relevant knowledge.
SeCoKD models also simplify complex tasks, demonstrating a higher capability to internalize and utilize fewer demonstrations. This benefit is quantified through metrics distinguishing positive and negative demonstrations and classifying task difficulty based on model responses.
Overall, SeCoKD offers a promising solution for enhancing LLM performance in few-shot and zero-shot learning contexts, providing a more efficient and scalable approach for leveraging demonstrations in language model training.
\section{Limitations}
While SeCoKD shows significant promise, there are several limitations to consider.
Firstly, the scope of our experiments is limited to models with fewer than 10 billion parameters due to computational constraints. This restriction may limit the generalizability of our findings to larger models, which are increasingly prevalent in current research and applications \citep{chung2022scaling, wei2023larger}.
Secondly, the benchmarks used in this study are focused primarily on reasoning tasks. While these benchmarks are diverse, extending the evaluation to include a broader range of tasks, such as language generation \citep{li2023towards}, summarization \citep{he2023icl}, or translation \citep{zhu2024towards}, would provide a more comprehensive understanding of SeCoKD's effectiveness. Moreover, more cross-studies would help to assess the sustainability of SeCoKD's performance improvements over different types of tasks.
Thirdly, in this study, we primarily focus on the self-KD settings, we save the opportunity to study distillation between different scales of models for future work.
Finally, the computational overhead associated with training using SeCoKD, especially in resource-constrained environments, needs further investigation. 
Addressing these limitations in future research will be essential for fully realizing the potential of SeCoKD and extending its applicability to a wider range of LLMs and tasks.


\bibliography{acl_latex}

\appendix

\section{Further Results on Few-Shot Learning}
\label{sec:appendix}

\label{sec:results}
\begin{figure}[htbp]
  \centering
  \includegraphics[width=\columnwidth]{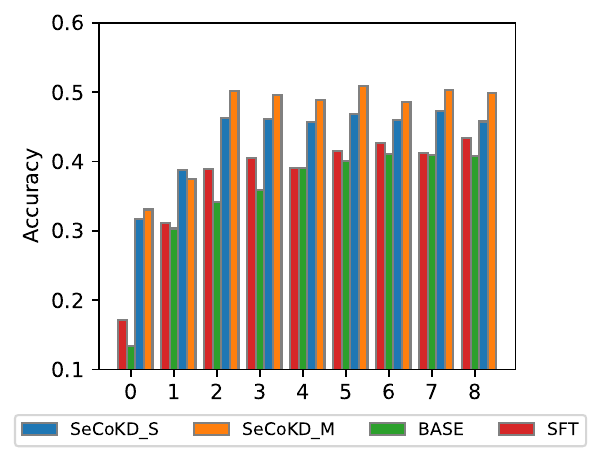}
  \caption {Comparison of 4 methods using Llama2 with different shot
numbers. The X-axis represents the number of demonstrations. The Y-axis shows the average accuracy of all six tasks. }
  \label{fig:llama2}
\end{figure}

\begin{figure}[htbp]
  \centering
  \includegraphics[width=\columnwidth]{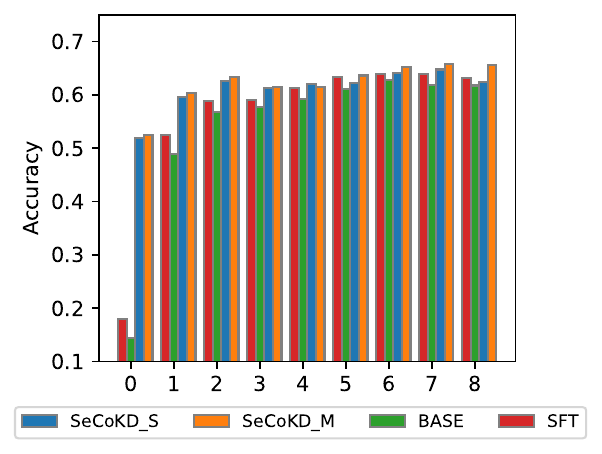}
  \caption {Comparison of 4 methods using Mistral with different shot
numbers. The X-axis represents the number of demonstrations. The Y-axis shows the average accuracy of all six tasks.}
  \label{fig:mistral}
\end{figure}

Figure \ref{fig:llama2} and Figure \ref{fig:mistral} show the average performance comparison using Llama2 7B and Mistral 7B models. For the Llama2 model, we see a huge improvement when training with SeCoKD. However, our main conclusion stays unchanged: compared to SFT and the base model, SeCoKD has a much better performance in the zero-shot and one-shot settings.

\section{Hyperparameters}
\label{sec:hyperparameters}
Table~\ref{tab:lora} summarizes the Lora configurations we used in our study. We used a relatively small rank (fewer trainable parameters) since we do not want to teach the model new associations beyond its knowledge. We target the 4 main linear layers of a transformer block but we did not tune this hyperparameter.

For training, we used the paged\_adamw\_32bit optimizer as suggested in the QLoRA paper\citep{dettmers2023qlora}. The batch size for training and evaluation is two because of the computational limitation. The learning rate is set to 1e-4 with a warmup ratio of 0.02. The best checkpoint evaluated on the testing set is saved as the final result.

\begin{table}[htpb]
\centering
\resizebox{\columnwidth}{!}{%
\begin{tabular}{ll}
\hline
r (rank)        & 32                                                 \\
lora\_alpha     & 64                                                 \\
target\_modules & {[} "q\_proj", "k\_proj", "out\_proj","v\_proj"{]} \\
lora\_dropout   & 0.05                                               \\
bias            & "none"                                             \\ \hline
\end{tabular}%
}
\caption{Lora configuration for all models.}
\label{tab:lora}
\end{table}

\section{Demonstration Pool}
\label{sec:demonstration}

We reuse the existing demonstrations for GSM8K, COIN-FLIP, and CSQA tasks from \citet{wei2022chain}. For SVAMP, we use the same set of demonstrations as for GSM8K since they are both mathematical reasoning tasks with similar formats. For ARC-C we present the curated demonstrations in Table~\ref{tab:arc prompt}.
\begin{table*}[htpb]
\centering
\begin{tabular}{ p{1\linewidth}}
\hline
\textbf{1.} Question:George wants to warm his hands quickly by rubbing them. Which skin surface will produce the most heat? (A) dry palms. (B) wet palms. (C) palms covered with oil. (D) palms covered with lotion. Answer: Dry surfaces will more likely cause more friction via rubbing than other smoother surfaces, hence dry palms will produce the most heat. The answer is: (A)\\ \hline
\textbf{2.} Question:Which factor will most likely cause a person to develop a fever? (A) a leg muscle relaxing after exercise. (B) a bacterial population in the bloodstream. (C) several viral particles on the skin. (D) carbohydrates being digested in the stomach.
Answer: Option (B), bacterial population is the most likely cause for a person developing fever. The answer is: (B)
 \\ \hline
\textbf{3.} Question:Which change in the state of water particles causes the particles to become arranged in a fixed position? (A) boiling. (B) melting. (C) freezing. (D) evaporating.
Answer: When water is freezed, the particles are arranged in a fixed position; the particles are still moving for all other options.  The answer is: (C)
 \\ \hline
\textbf{4.} Question:When a switch is used in an electrical circuit, the switch can (A) cause the charge to build. (B) increase and decrease the voltage. (C) cause the current to change direction. (D) stop and start the flow of current.
Answer: The function of a switch is to start and stop the flow of a current. The answer is: (D)
 \\ \hline
\textbf{5.} Question:Which of the following statements best explains why magnets usually stick to a refrigerator door? (A) The refrigerator door is smooth. (B) The refrigerator door contains iron. (C) The refrigerator door is a good conductor. (D) The refrigerator door has electric wires in it.
Answer: Since iron is a ferromagnetic material that is strongly attracted to magnets The answer is: (B)
 \\ \hline
\textbf{6.} Question:Which of these do scientists offer as the most recent explanation as to why many plants and animals died out at the end of the Mesozoic era? (A) worldwide disease. (B) global mountain building. (C) rise of mammals that preyed upon plants and animals. (D) impact of an asteroid created dust that blocked the sunlight.
Answer: The most accepted and supported explanation among scientists for the mass extinction event at the end of the Mesozoic era is (D) the impact of an asteroid that created dust blocking sunlight. This event led to drastic changes in climate and ecosystems, making it impossible for many species to survive. The answer is: (D)
 \\ \hline
\textbf{7.} Question:A boat is acted on by a river current flowing north and by wind blowing on its sails. The boat travels northeast. In which direction is the wind most likely applying force to the sails of the boat? (A) west. (B) east. (C) north. (D) south.
Answer: The boat travels northeast, and the river current flows north. This implies that to achieve a northeast direction, the boat must receive an additional force component to the east. The answer is: (B)
 \\ \hline
\textbf{8.} Question:Which landform is the result of the constructive force of a glacier? (A) valleys carved by a moving glacier. (B) piles of rocks deposited by a melting glacier. (C) grooves created on a granite surface by a glacier. (D) bedrock hills roughened by the passing of a glacier.
Answer: The constructive process results in the accumulation of debris and rocks, contributing to the formation of new landforms such as moraines, which are essentially piles of rocks and soil deposited by glaciers. The answer is: (B)
 \\ \hline
\end{tabular}
\caption{Full prompts for the ARC-C dataset.}
\label{tab:arc prompt}
\end{table*}

\section{Rationale for SFT training}
\label{sec:rationale}
For the datasets \textbf{GSM8K} and \textbf{AQUA-RAT} the creators provide rationales for each sample. For the rest, we prompt GPT3.5 with query and the answer to generate the corresponding rationales. The template is: "The answer to the question \{query\} is \{answer\}. Please generate a short rationale to justify the answer."

\begin{acronym}
    \acro{ICL}[ICL]{In-Context Learning}
    \acro{PLM}[PLM]{Pretrained Language Model}
    \acro{KD}[KD]{Knowledge Distillation}
    \acro{LLM}{Large Language Model}
    \acro{CoT}{Chain-of-Thought}
    \acro{SFT}{Supervised Fine-tuning}
    \acro{LLMs}{Large Language Models}
\end{acronym}

\end{document}